\documentclass[proceed]{article} \usepackage{proceed2e}
\usepackage{times, amsmath, amssymb, graphicx, breqn}
\usepackage[algoruled]{algorithm2e} \usepackage[round]{natbib}
\usepackage{subcaption} \usepackage{xcolor} \title{Importance Sampled
  Stochastic Optimization for Variational Inference}
\author{ {\bf Joseph Sakaya} \\
  Helsinki Institute for Information Technology~HIIT\\
  Department of Computer Science \\
  University of Helsinki, Finland\\
  \And
  {\bf Arto Klami}  \\
  Helsinki Institute for Information Technology~HIIT\\
  Department of Computer Science \\
  University of Helsinki, Finland\\
}

\DeclareMathOperator*{\Softplus}{Softplus}
 
\DeclareMathOperator*{\Det}{det} 
\begin{document}
\maketitle
\begin{abstract}
  Variational inference approximates the posterior distribution of a
  probabilistic model with a parameterized density by maximizing a
  lower bound for the model evidence. Modern solutions fit a flexible
  approximation with stochastic gradient descent, using Monte Carlo
  approximation for the gradients. This enables variational inference
  for arbitrary differentiable probabilistic models, and consequently
  makes variational inference feasible for probabilistic programming
  languages. In this work we develop more efficient inference
  algorithms for the task by considering importance sampling estimates
  for the gradients. We show how the gradient with respect to the
  approximation parameters can often be evaluated efficiently without
  needing to re-compute gradients of the model itself, and then
  proceed to derive practical algorithms that use importance sampled
  estimates to speed up computation.  We present importance sampled
  stochastic gradient descent that outperforms standard stochastic
  gradient descent by a clear margin for a range of models, and
  provide a justifiable variant of stochastic average gradients for
  variational inference.
\end{abstract}

\section{INTRODUCTION}

Variational inference considers parametric approximations for
posterior densities of probabilistic models. Following
\citet{Jordan1999} the classical variational approximation algorithms
are based on coordinate descent algorithms for which individual steps
of the algorithm are often carried out analytically. This limits the
use of variational approximation to models with conjugate priors (or
simple extensions of those) and restricts the family of potential
approximating distributions based on analytic tractability.

Recent advances in variational approximation have lead to a phase
transition; instead of closed-form updates, the approximation is
nowadays often fit using generic gradient descent techniques instead
-- for a good overview see, {\it e.g.}, \citet{gal}.  The key behind
these advances is in using Monte Carlo approximation to estimate the
gradient of the objective function that is an integral over the
approximating distribution.  This can be done in two alternative
ways. The \emph{reparameterization estimate} \citep{dsvi, vae,
  salimans} allows expressing the gradient of the objective function
using the gradients of the model itself, whereas the \emph{score
  function estimate} \citep{bbvi} is based on gradients of the
approximation.  Given the new family of algorithms we can apply
variational approximation for a considerably wider range of
probabilistic models, enabling for example use of variational
inference as the inference backend in probabilistic programming
languages \citep{advi,pymc3,edward}.

The main research efforts in variational inference are nowadays geared
towards making the approach applicable to a still wider family of
models, by constructing even more flexible approximations
\citep{rezende15, ranganath16, kingma16} or by generalizing the
gradient estimators \citep{grep, naesseth}.  The question of how
exactly the resulting optimization problem is solved has largely
remained unattended to -- practically all authors are satisfied with
standard stochastic gradient descent (SGD) as the underlying
optimizer, although some effort has been put into improving
convergence by reducing the variance of the gradient estimate near the
optimum \citep{roeder17}.

We turn our attention to the optimizer itself, looking into ways of
speeding up the computation of gradient-based variational
approximation. Practically all of the computational effort during
learning goes into evaluating the gradient of the model (or the
approximation if using the score-function estimate). Our contribution
is in reducing the number of times we need to evaluate the gradient of
the model during the optimization process, based on an importance
sampling scheme specifically designed for optimization problems where
the gradients are computed using Monte Carlo approximation.

The key observation is that the gradient of the objective function
with respect to the parameters of the approximation consists of two
parts. One part is the gradient of the model itself, evaluated at
parameter values drawn from the approximation, whereas the other part
is the gradient of the transformation used in the reparameterization
estimate. We show that the gradient required for optimization can be
computed for the newly updated approximation without re-computing the
first part, which is computationally heavier. Instead, we can re-use
existing computation by appropriately modifying and re-weighting the
available terms.

We show how to formulate this idea in a justified manner, by
constructing an importance sampling estimate for the gradient.
Importance sampling is typically used for cases where one cannot
sample from the distribution of interest but instead has to resort to
sampling from a related proposal distribution. In our case we could
sample from the distribution of interest -- the current approximation
-- but choose not to, since by using an earlier approximation as a
proposal we can avoid costly computation. The idea is conceptually
similar to the way \citet{ep} reuses samples from previous iterations
in expectation propagation.

Since the advances in our case are related to the computation of the
gradient itself, the idea can readily be combined with several
optimization algorithms.  In this work, we derive practical algorithms
extending standard SGD and stochastic average gradients
\citep{sage}. We demonstrate them in learning a variational
approximation for several probabilistic models, showing how they
improve the convergence speed in a model-independent manner.

In the following we first give a brief overview of the
state-of-the-art in gradient-based variational approximation, covering
both the gradient estimates and stochastic optimization algorithms in
Section~2.  We then proceed to describe the importance sampling
estimate for the gradient in Section~3, followed by practical
optimization algorithms outlined in Section~4. Empirical experiments
and illustrations are provided in Section~5.

\section{BACKGROUND}

\subsection{VARIATIONAL APPROXIMATION}

Variational inference refers to approximating the posterior
distribution $p(z|x)$ of a probabilistic model $p(x,z)$ using a
distribution $q_\lambda(z)$ parameterized by $\lambda$.  Usually this
is achieved by maximizing a lower bound $\mathcal{L}(\lambda)$ for the
evidence (also called the marginal likelihood) $p(x)$:
\begin{equation}
  \mathcal{L}(\lambda) = \int q_\lambda(z) \log \frac{p(x, z)} {q_\lambda(z)}\ \delta z \le p(x).
  \label{eq:ELBO}
\end{equation}
Traditionally, the problem has been made tractable by assuming a
factorized mean-field approximation \mbox{$q_\lambda(z) = \prod_i
  q_{\lambda_i}(z_i)$} and models with conjugate priors, resulting in
closed-form coordinate ascent algorithms specific for individual
models -- for a full derivation and examples, see, {\it e.g.},
\citet{blei16}.

In recent years several novel types of algorithms applicable for a
wider range of models have been proposed \citep{dsvi, vae, salimans,
  bbvi}, based on direct gradient-based optimization of the lower
bound \eqref{eq:ELBO}.  The core idea behind these algorithms is in
using Monte Carlo estimates for the loss and its gradient
\begin{equation}
  \nabla_{\lambda} \mathcal{L}(\lambda) = 
  \nabla_\lambda \mathbb{E}_{q_\lambda(z)}[\log p(x,z) - \log q_\lambda(z)].
  \label{eq:ELBOgrad}
\end{equation}
Given such estimates, the inference problem can be solved by standard
gradient descent algorithms. This enables inference for non-conjugate
likelihoods and for complex models for which closed-form updates would
be hard to derive, making variational inference a feasible inference
strategy for probabilistic programming languages \citep{advi, edward,
  pymc3}.  In the following, we briefly describe two alternative
strategies of estimating the gradient.  In Section \ref{sec:method} we
will then show how the proposed importance sampling technique is
applied for both cases.

\subsubsection{REPARAMETERIZATION ESTIMATE}

The reparameterization estimate for \eqref{eq:ELBO} (and consequently
\eqref{eq:ELBOgrad}) is based on representing the approximation
$q_\lambda(z)$ using a differentiable transformation $z = f(\epsilon,
\lambda)$ of an underlying standard distribution $\phi(\epsilon)$ that
does not have any free parameters. The core idea of how this enables
computing the gradient was developed simultaneously by \citet{vae,
  salimans} and \citet{dsvi} with many of the mathematical details
visible already in the early work by \citet{opper}.  Plugging the
transformation $f(\cdot)$ into \eqref{eq:ELBO} gives
\[
\mathcal{L}(\lambda) = \int \phi(\epsilon) \log \frac{p(x,
  f(\epsilon,\lambda))|\Det_{J_f}(\epsilon,
  \lambda)|}{\phi(\epsilon)}\ \delta \epsilon,
\]
where the integral is over the standard distribution that does not
depend on $z$, and $|\Det_{J_f}(\epsilon, \lambda)|$ is the absolute
value of the determinant of the Jacobian of $f(\epsilon, \lambda)$.
Consequently, it can be replaced by a stochastic approximation
\[
\mathcal{L}(\lambda) = \frac{1}{M} \sum_{m=1}^M \log \frac{p(x,
  f(\epsilon_m,\lambda))|\Det_{J_f}(\epsilon_m,
  \lambda)|}{\phi(\epsilon_m)},
\]
where $\epsilon_m$ is drawn from $\phi(\epsilon)$.  We can now easily
compute the gradients using the chain rule, by first differentiating
$\log p(x, f(\epsilon, \lambda))$ w.r.t $z$ and then $z = f(\epsilon,
\lambda)$ w.r.t $\lambda$, resulting in \vspace{-1em}
\begin{dmath}
  \nabla_{\lambda} \mathcal{L}(\lambda) \approx \frac{1}{M}
  \sum_{m=1}^M \left[\nabla_{z} \log p(x, z_m)
    \nabla_{\lambda}f(\epsilon_m, \lambda) +
    \nabla_{\lambda}|\text{det}_{J_f}(\epsilon_m, \lambda)|\right].
  \label{eq:MCELBO}
\end{dmath}

The combination of the standard distribution $\phi(\epsilon)$ and the
transformation $f(\epsilon, \lambda)$ defines the approximation
family. For example, $\phi(\epsilon) = \mathcal{N}(0, I)$ and
$f(\epsilon, \lambda=\{\mu, L\}) = \mu + L \epsilon$ defines arbitrary
Gaussian approximations \citep{opper, dsvi}, where $L$ is the Cholesky
factor of the covariance.  To create richer approximations we can
concatenate multiple transformations; \citet{advi} combines the
transformation above with a rich family of univariate transformations
designed for different kinds of parameter constraints. For example, by
using $f(\epsilon, \lambda=\{\mu, L\}) = \Softplus(\mu + L \epsilon)$
we can approximate parameters constrained for positive
values. Alternatively, we can directly reparameterize other common
distributions such as Gamma or Dirichlet -- see \citet{naesseth, grep}
for details.

\subsubsection{SCORE FUNCTION ESTIMATE}

An alternative estimate for \eqref{eq:ELBOgrad} can be derived based
on manipulation of the log-derivatives; the use of the estimate for
simulation of models was originally presented by \citet{score} and its
use for variational approximation by \citet{bbvi}. The estimate for
$\nabla_\lambda \mathcal{L}(\lambda) $ is provided by
\begin{equation*}
  \mathbb{E}_{q_\lambda(z)}[(\log p(x,z) - \log q_\lambda(z)) \nabla_\lambda \log q_\lambda(z)],
\end{equation*}
which again leads into a straightforward Monte Carlo approximation of
$\nabla_\lambda \mathcal{L}(\lambda)$ as
\begin{equation}
  \frac 1M \sum_{m = 1}^M(\log p(x,z_m) - \log q_\lambda(z_m)) \nabla_\lambda \log q_\lambda(z_m),
  \label{eq:score}
\end{equation}
where $z_m \sim {q_\lambda(z)}$.  The notable property of this
technique is that it does not require derivatives of the model (that
is, $\log p(x, z)$) itself, but instead relies solely on derivatives
of the approximation. This is both a pro and a con; the model does not
need to be differentiable, but at the same time the estimate is not
using the valuable information the model gradient provides. This is
shown to result in considerably higher variance compared to the
reparameterization estimate, often by orders of magnitude.  Variance
reduction techniques \citep{bbvi} help, but for differentiable models
the reparameterization technique is typically considerably more
efficient \citep{naesseth,grep}.

\subsection{STOCHASTIC GRADIENT OPTIMIZATION}

Given estimates for the gradient \eqref{eq:ELBOgrad} computed with
either method, the optimization problem\footnote{Here cast as
  minimization of negative evidence, to maintain consistent
  terminology with gradient descent literature} is solved by standard
gradient descent techniques. In practice all of the automatic
variational inference papers have resorted to stochastic gradient
descent (SGD) on mini-batches, adaptively tuning the step lengths with
the state-of-the-art techniques.

In recent years several more advanced stochastic optimization
algorithms have been proposed, such as stochastic average gradients
(SAG) \citep{sage}, stochastic variance reduced gradients (SVRG)
\citep{svrg}, and SAGA that combines elements of both \citep{saga}.
However, to our knowledge these techniques have not been successfully
adapted for automatic variational inference. In Section \ref{sec:isag}
we will present a new variant of SAG that works also when the
gradients are estimated as Monte Carlo approximations, and therefore
briefly describe below the basic idea behind stochastic average
gradients.

SAG performs gradient updates based on an estimate for the full batch
gradient, obtained by summing up gradients stored for individual data
points (or for mini-batches to save memory). Whenever a data point is
seen again during the optimization the stored gradient for that point
is replaced by the gradient evaluated at the current parameter
values. The full gradient estimate hence consists of individual
gradients estimated for different parameter values; the most recently
computed gradients are accurate but the ones computed long time ago
may correspond to vastly different parameter values. This introduces
bias \citep{saga}, but especially towards the convergence the variance
of the estimated full batch gradient is considerably smaller than that
of the latest mini-batch, speeding up convergence.

\section{METHOD}
\label{sec:method}
\begin{figure}[t!]
  \centering
  \includegraphics[scale = .6]{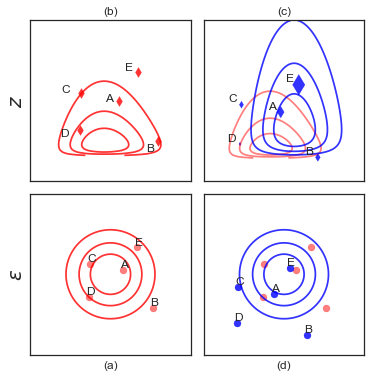}
  \caption{\textbf{Importance sampling illustration for
      reparametrization gradients}: A Monte Carlo estimate for the
    model evidence lower bound is obtained by drawing $M$ (here $5$)
    samples $z_m$ from the approximation $q(\lambda)$ depicted by the
    contours in top left panel. The reparameterization estimate does
    this by actually drawing the samples $\epsilon_m$ from a standard
    distribution depicted in the bottom left panel, transforming them
    to the parameter space with $f(\epsilon, \lambda)$.  The
    importance sampling estimate for the evidence lower bound of
    another approximation $q(\lambda')$ (top right) is obtained by
    keeping $z_m$ fixed and computing the $\epsilon_m' = f^{-1}(z_m,
    \lambda')$ that would have been required as draws from the
    standard distribution (bottom right) to produce them. The estimate
    then re-weights the already computed log-likelihoods $\log p(x,
    z_m)$ with $w_m = \frac{\phi(\epsilon_m')}{\phi(\epsilon_m)}$ to
    approximate the evidence lower bound; here sample E has higher
    probability under the new approximation, shown as larger diamond,
    and sample D has lower probability.  The same basic procedure
    applies also for estimating the gradients instead of the
    evidence.}
  \label{fig:normal}
\end{figure}

All gradient-based optimization algorithms follow the same basic
pattern of computing a gradient for a mini-batch of samples and
updating the parameters. The computational effort required goes almost
solely into evaluating the gradient of the loss. To speed up the
optimization, we next present a technique that allows computationally
lighter evaluation of the gradient in scenarios where the gradient is
computed using a Monte Carlo approximation.  The presentation here is
based on the reparameterization estimate \eqref{eq:MCELBO} that
benefits more off this treatment -- as will become evident later --
but for completeness we discuss also the score function estimate
\eqref{eq:score} in Section~\ref{sec:score}.

The Monte Carlo approximation for estimating the gradient depends on
the data $x$ and a set of $M$ parameters $z_m$ drawn from the current
approximation. As highlighted in \eqref{eq:MCELBO}, the actual
computation factorizes into $\nabla_z p(x, z_m)$ that depends only on
$x$ and $z_m$ and into $\nabla_\lambda f(\epsilon, \lambda)$ and
$\nabla_{\lambda}|\text{det}_{J_f}(\epsilon_m, \lambda)|$ that depend
only on $\epsilon_m$ and $\lambda$. The former part is typically
considerably more computationally expensive. The observation that the
slower part does not directly depend on the parameters hints that it
should be possible to avoid re-computing the term even if the
approximation changes, and this indeed is the case as explained next.

Assume we have already estimated the gradient at some parameters
$\lambda$, implying that we have also evaluated $\nabla_z p(x, z_m)$
for some set of $z_m$.  The question now is how to estimate the
gradient at parameters $\lambda' = \lambda + \delta$ that are
(typically only slightly) different. It turns out this can be done
using the well-known concept of \emph{importance sampling} originally
designed for approximating expectations when we cannot directly draw
samples from the density of interest.  In our case, however, we
\emph{could} draw samples directly from the new approximation, but
\emph{choose not to} since the gradient can be estimated also using
the old approximation as a proposal distribution. That is, we are
using importance sampling for an unusual reason but can still use all
the standard tools.

Typically, we use importance sampling to find the expectation of a
function $f(z)$ over the target distribution $p(z)$ if we are able to
draw samples only from a proposal distribution $q(z)$. The expectation
of $f(z)$ over $p(z)$ can then be approximated by
\parbox{\linewidth}{
\begin{equation}
\hspace{-.2em} 
\mathbb{E}[f] = \int \frac{p(z)}{q(z)}
  f(z) q(z) \ \delta z = \frac 1 M \sum_{m= 1}^M
  \frac{p(z_m)}{q(z_m)} f(z_m).
  \label{eq:is}
\end{equation}
}
The quantities $w_m = {p(z_m)}/ {q(z_m)}$ are the \emph{importance
  weights} that correct the bias introduced by sampling from the wrong
distribution $q(z)$ \citep{bishop}. The weights $w_m$ are non-negative
and tend to zero when $p(z)$ is completely mismatched to $q(z)$, and
$w_m > 1$ when the sample $z_m$ is more likely under the $p(z)$.  The
estimate above is unbiased, but has high -- potentially infinite --
variance when $q(z)$ and $p(z)$ are dissimilar. Next we show how importance sampling 
can be used for evaluating the reparameterization gradient \eqref{eq:MCELBO}. 

To save computation we want to re-use the model gradients $\nabla_z
\log p(x, z_m)$ already available for certain values of $z_m$ and
hence need to consider estimates that keep these values fixed.  This
means we need to find the $\epsilon_m'$ under the new approximation
$q_{\lambda'}(z)$ that correspond to these values, by computing
$\epsilon_m' = f^{-1}(z_m, \lambda')$. Given these values we can
evaluate the necessary quantities to compute both the importance
weights and the other terms ($\nabla_{\lambda} f(\epsilon_{m}',
\lambda')$ and $\nabla_{\lambda}|\text{det}_{J_f}(\epsilon_m,
\lambda)|$) required for evaluating the gradient itself.

The resulting importance sampling estimate for \eqref{eq:MCELBO} is
\begin{dmath}
  \nabla^i_{\lambda} \mathcal{L}(\lambda) \approx \frac{1}{M}
  \sum_{m=1}^M \left[{\nabla_{z} \log p(x, z_m)
    }\nabla_{\lambda}f(\epsilon'_m, \lambda') +
    \nabla_{\lambda}|\text{det}_{J_f}(\epsilon'_m, \lambda')|\right]
  \times w_m,
  \label{eq:impgrad}
\end{dmath}
where the $i$ in $\nabla^i_{\lambda} \mathcal{L}(\lambda) $ refers to
the importance-sampled estimate of the gradient. The computationally
expensive part of the gradient $\nabla_{z} \log p(x, z_m)$ is already
available and need not be computed. The rest of the terms are
efficient to evaluate, and hence the whole gradient estimate is
obtained in a fraction of a time compared to computing it from
scratch. The importance weights are provided by
\begin{equation}
  w_m = 
  \frac{\phi(\epsilon_m')}{\phi(\epsilon_m)} \label{eq:weight}
\end{equation}
and hence only require evaluating densities of the standard
distribution underlying the approximation.  The above description is
summarized in Algorithm~\ref{alg:impgrad} and illustrated graphically
in Figure~\ref{fig:normal}.

It is also worth noting that if $f(\cdot)$ is constructed as a series
of transformations, for example as element-wise transformations of
reparameterized normal distribution as done by \citep{advi}, then
parts of the term $\nabla_{\lambda}|\text{det}_{J_f}(\epsilon'_m,
\lambda')|$ can (and naturally should) also be re-used. This is,
however, of secondary importance compared to the computational saving
of re-using $\nabla_z \log p(x, z_m)$.

A practical challenge with importance sampling is that for
high-dimensional densities the weights $w_m$ easily tend to zero.
Variational approximation is, however, often conducted for
approximations that factorize over the parameters of the approximation
as $q(\lambda) = \prod_{s} q(\lambda_{S_s})$, where $\{S_s\}$ is a
partitioning of the parameter vector.  The importance sampling
estimate can -- and should -- be done for each factor separately,
since the gradient $\nabla_z \log p(x, z_m)$ can be computed for each
approximation factor independently even if $\log p(x, z_m)$ itself
does not factorize.  We show in Section~\ref{sec:dimensionality} that
the technique helps at least until factors consisting of roughly ten
parameters.

\IncMargin{1em}
\begin{algorithm}[t]
  \DontPrintSemicolon
  \SetKwData{Left}{left}\SetKwData{This}{this}\SetKwData{Up}{up}
  \SetKwFunction{Union}{Union}\SetKwFunction{FindCompress}{FindCompress}
  \SetKwInOut{Input}{input}\SetKwInOut{Output}{output} \Input{Samples
    $z_m$ and $\epsilon_m$, gradient $\nabla_zp(x,z_m)$,
    transformation $f()$, current approximation $\lambda'$ }
  \Output{Importance sampled gradients $\nabla_\lambda^i
    \mathcal{L}(\lambda)$} \BlankLine $\epsilon_m' \leftarrow
  f^{-1}(z_m, \lambda')$\; $w_m \leftarrow
  \frac{\phi(\epsilon'_m)}{\phi(\epsilon_m)}$\; Calculate
  $\nabla^i_{\lambda} \mathcal{L}(\lambda)$ using \eqref{eq:impgrad}
  \;
  \caption{Importance sampled gradients}
  \label{alg:impgrad}
\end{algorithm}\DecMargin{1em}

\subsection{REPARAMETERIZATION GRADIENT EXAMPLE}

To further clarify the derivation above, we next illustrate the
procedure for the common scenario of Gaussian reparameterization
combining $\phi(\epsilon) = \mathcal{N}(0, I)$ with $z = \mu + L
\epsilon$, denoting $\lambda = \{\mu, L\}$.  Given a set of $z_m$
drawn from $q(\lambda)$ we want to estimate the gradient evaluated at
$\lambda'$ using \eqref{eq:impgrad}.

First we compute $\epsilon_m' = f^{-1}(z_m, \lambda') = L^{-1}(z_m -
\mu)$ and evaluate $\phi(\epsilon_m')$ under the standard normal
distribution. The weight $w_m$ can then readily be evaluated using
\eqref{eq:weight}, computing $\phi(\epsilon_m)$ as well if it has not
already been evaluated because of being used for another importance
sampling estimate.

To compute the gradient \eqref{eq:impgrad} we need to re-compute
$\nabla_{\lambda} f(\epsilon_m', \lambda')$ and
$\nabla_{\lambda}|\text{det}_{J_f}(\epsilon'_m, \lambda')|$.  For
$\mu$ these terms are simply identity and zero, whereas for $L$ we get
$z_m$ and $\nabla_{L} \log |L'| \equiv \Delta_{L'}$. The exact form of
$\Delta_{L'}$ depends on the assumptions made for $L$; see
\citet{dsvi} for details.  The final importance sampled estimate for
the gradient is then
\begin{align*}
  \nabla_{\mu}^{i} \mathcal{L}(\mu', L') &\approx
  \frac{1}{M} \sum_{m=1}^M w_m \nabla_{z} \log p(x, z_m)\\
  \nabla_{L}^{i} \mathcal{L}(\mu', L') &\approx \frac{1}{M}
  \sum_{m=1}^M w_m \left [ \nabla_{z} \log p(x, z_m)
    \epsilon'_m %\times (z_m - \mu')^T L^{-T}
    + \Delta_{L'} \right ]. \notag
\end{align*}

An important observation here is that the importance sampling
procedure does not merely re-weight the terms $\nabla_z \log p(x,
z_m)$, but in addition the transformation that converts them into the
$\lambda$ space changes because of the new values of $\epsilon_m'$.
These values depend on the approximation parameters in a non-linear
fashion and hence the gradient itself is a non-linear transformation
of the gradient evaluated at $\lambda$ (for a graphical illustration,
see Figure~\ref{fig:normalcontour}).  This is crucially important for
development of the practical optimization algorithms in
Section~\ref{sec:algorithms}; if the transformation was linear then
the importance sampling estimate would not necessarily provide
improvement over careful adaptation of element-wise learning rates.

\subsection{SCORE FUNCTION ESTIMATE}
\label{sec:score}
Above we discussed the importance sampling estimate from the
perspective of the reparameterization estimate. For completeness we
also show how it can be applied for the score function estimate, and
discuss why it is less useful there.

The Monte Carlo approximation for the score function \eqref{eq:score}
is obtained directly by drawing samples from the approximation
$q(\lambda)$. To approximate the gradient for $q(\lambda')$ we merely
use the standard importance sampling equation \eqref{eq:is} to obtain
the approximation
\[
\frac{1}{M} \sum_{m=1}^M w_m \left[\log p(x, z_m) - \log
  q_\lambda(z_m)\right] \nabla_{\lambda} \log q_\lambda(z_m),
\]
where $w_m = \frac{q(z|\lambda')}{q(z|\lambda)}$. This is still an
unbiased estimate, but the computational saving is typically smaller
than in the reparameterization case. We do not need to evaluate $\log
p(x, z_m)$ since the samples $z_m$ are kept constant, but all other
terms need to be computed again and evaluating the gradient of the
approximation is not cheap. This estimate is only useful when the
evaluation of the log probability utterly dominates the total
computation.

\section{ALGORITHMS}
\label{sec:algorithms}
In the following we describe example optimization algorithms based on
the importance sampling idea. The details are provided for a
straightforward variant of SGD and for a generalization of stochastic
average gradients, but other related algorithms could be instantiated
as well.

\subsection{IMPORTANCE SAMPLED SGD}

\IncMargin{1em}
\begin{algorithm}[t]
  \DontPrintSemicolon
  \SetKwData{Left}{left}\SetKwData{This}{this}\SetKwData{Up}{up}
  \SetKwFunction{Union}{Union}\SetKwFunction{FindCompress}{FindCompress}
  \SetKwInOut{Input}{input}\SetKwInOut{Output}{output} \Input{data
    $X$, model $p(x,z)$, the variational approximation $q_\lambda(z)$,
    threshold $t$} \Output{variational parameters $\lambda^*$}
  \BlankLine \While{$\mathcal{L}(\lambda)$ has not converged}{
    \eIf{$\text{random\_uniform} < t$}{ Retrieve last stored $z_m$,
      $\epsilon_m$\ and $\nabla_z\log p(x,z_m)$\; Update
      $\nabla_\lambda\mathcal{L}$ using Algorithm
      \eqref{alg:impgrad}\; }{ Draw mini-batch $x$ from $X$\;
      $\epsilon_m \sim \phi(\epsilon)$\; $z_m \leftarrow f(\epsilon_m,
      \lambda)$\; Update $\nabla_\lambda\mathcal{L}$ using Equation
      \eqref{eq:MCELBO}\; Store $z_m$, $\epsilon_m$ and $\nabla_z\log
      p(x,z_m)$\; } Update $\lambda \leftarrow \lambda + \rho
    \nabla_\lambda \mathcal{L}(\lambda)$\; }
  \caption{Importance sampled SGD}
  \label{alg:isgd}
\end{algorithm}\DecMargin{1em}
Stochastic gradient descent estimates the gradient based on a
mini-batch and then takes a step along the gradient direction,
typically using adaptive learning rates such as those by \citet{adam,
  adagrad}.

The importance sampled SGD (I-SGD; Algorithm~\ref{alg:isgd}) follows
otherwise the same pattern, but for each mini-batch we conduct several
gradient steps instead of just one.  For the first one we evaluate the
gradient directly using \eqref{eq:ELBOgrad}. After updating the
approximation we apply Algorithm~\ref{alg:impgrad} to obtain an
importance-sampled estimate for the gradient evaluated at the new
parameter values, and proceed to take another gradient step using that
estimate.  For each step we use a proper estimate for the mini-batch
gradient that can, after the first evaluation, be computed in a
fraction of a time. After taking a few steps we then proceed to
analyze a new mini-batch, again needing to compute the gradient from
scratch since now $x$ has changed.

After passing through the whole data we have evaluated $\log p(x, z)$
and its gradient once for every data point, just as in standard
SGD. However, we have taken considerably more gradient steps, possibly
by a factor of ten.  Alternatively, we can think of it as performing
more updates given a constant number of model gradient evaluations.

A practical detail concerns the choice of how many steps to take for
each mini-batch. This choice is governed by two aspects. On one hand
we should not use the importance sampled estimate if the approximation
has changed too much since computing the $\nabla_z \log p(x, z_m)$
terms, recognized typically as $w_m$ tending to zero.  On the other
hand, we should not take too many steps even if the approximation does
not change dramatically, since the gradient is still based on just a
single mini-batch.

The empirical experiments in this paper are run with a simple
heuristic that randomly determines whether to take another step with
the current mini-batch or to proceed to the next one. This introduces
a single tuning parameter $t$ that controls the expected number of
steps per mini-batch. The algorithm is robust for this choice; we
obtain practical speedups with values ranging from $t = 0.5$ to
$t=0.9$.  Finally, importance-sampling could in principle result in
very large gradients if $w_m \gg 1$ for some $m$; we never encountered
this in practice, but a safe choice is to proceed to the next batch if
that happens.

\begin{figure}[t!]
  \centering
  \includegraphics[scale = .7]{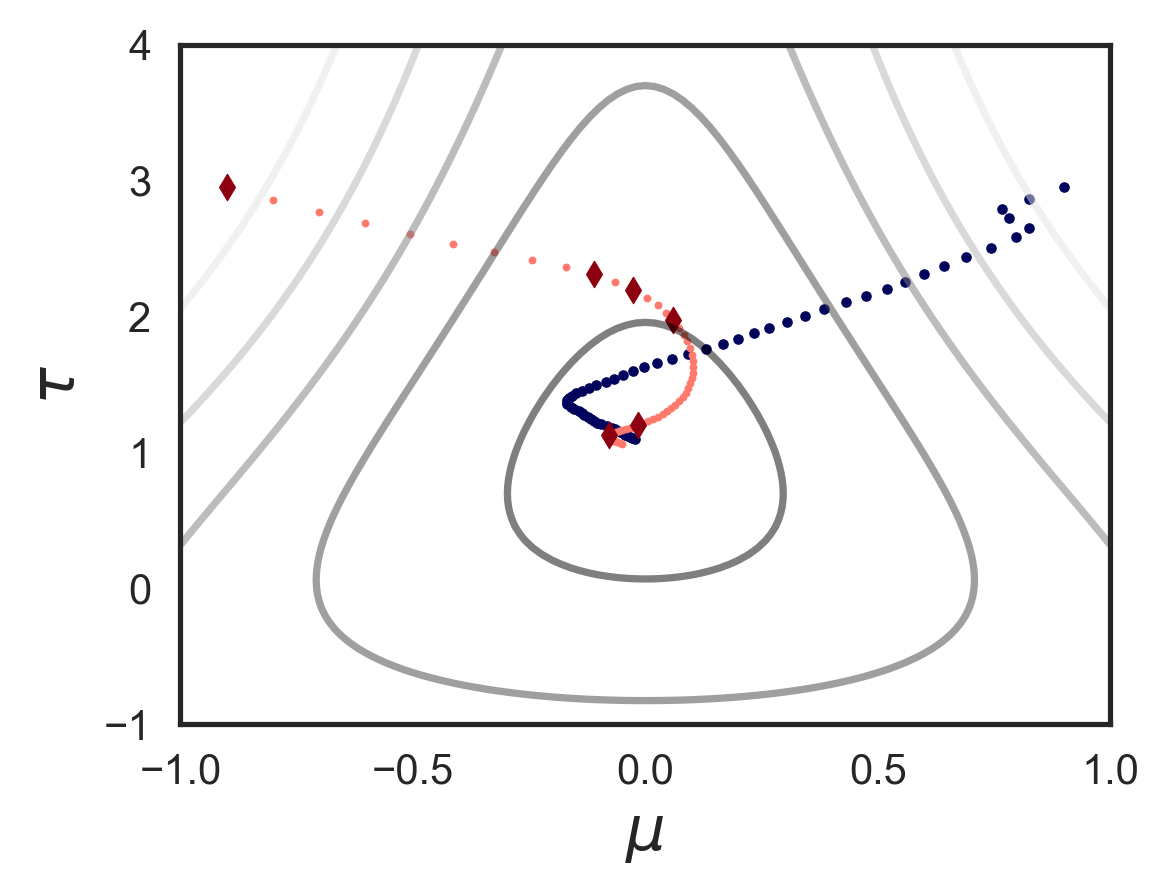}
  \caption{\textbf{Illustration of I-SGD.} I-SGD learns to approximate
    the posterior of a univariate normal distribution (parameterized
    by mean and precision) by computing the gradient of the evidence
    lower bound only a few times from scratch, indicated by red
    diamonds. Between these operations it performs multiple gradient
    steps (small red dots) using importance sampling estimates for the
    gradient that can be computed in a fraction of a time, while still
    following a non-linear trajectory between the red diamonds.  For
    comparison, standard SGD, needs to estimate the full gradient for
    every step (blue dots).}
  \label{fig:normalcontour}
\end{figure}

For a practical illustration of the algorithm, see
Figure~\ref{fig:normalcontour} that approximates the posterior over
the mean and precision of a normal model. Here $t=0.9$ and hence we
take on average $9$ importance-sampled gradient steps for each
mini-batch. The I-SGD algorithm reaches the optimum in roughly as many
steps as conventional SGD but achieves it almost ten times faster.

\IncMargin{1em}
\begin{algorithm}[t!]
  \DontPrintSemicolon
  \SetKwData{Left}{left}\SetKwData{This}{this}\SetKwData{Up}{up}
  \SetKwFunction{Union}{Union}\SetKwFunction{FindCompress}{FindCompress}
  \SetKwInOut{Input}{input}\SetKwInOut{Output}{output} \Input{data
    $X$, model $p(x,z)$, the variational approximation $q_\lambda(z)$,
    batches $B$} \Output{variational parameters $\lambda^*$}
  \BlankLine \While{$\mathcal{L}(\lambda)$ has not converged}{
    \ForEach{mini-batch $x_b$ in $X$}{ $\epsilon^b_m \sim
      \phi(\epsilon)$\; $z^b_m \leftarrow f(\epsilon_m, \lambda)$\;
      Calculate $\nabla^b_\lambda\mathcal{L}$ using Equation
      \eqref{eq:MCELBO}\; Store $z^b_m$, $\epsilon^b_m$ and
      $\nabla^b_z\log p(x_b,z_m)$\; \ForEach{mini-batch $x_c$ in
        $X\backslash\{x_b\}$}{ Retrieve $z^c_m$, $\epsilon^c_m$ and
        $\nabla^c_z\log p(x_c,z_m)$\; Update
        $\nabla^c_\lambda\mathcal{L}$ using Algorithm
        \eqref{alg:impgrad}\; } $\nabla_\lambda\mathcal{L} \leftarrow
      \sum_{n = 1}^{B} \nabla_\lambda^n\mathcal{L}$\; Update $\lambda
      \leftarrow \lambda + \rho \nabla_\lambda \mathcal{L}(\lambda)$\;
    } }
  \caption{Importance sampled SAG}
  \label{alg:isag}
\end{algorithm}\DecMargin{1em}

\subsection{IMPORTANCE SAMPLED SAG}
\label{sec:isag}
Stochastic average gradients \citep{sage} stores the batch gradient
and iteratively updates it for the samples in a given mini-batch.  In
the following we derive a variant of SAG (Algorithm~\ref{alg:isag})
that uses importance sampling to both re-weight and update the
gradients for the historical mini-batches using \eqref{eq:impgrad},
helping to detect and avoid using stale gradients whose parameter
values have changed so much since computing them.

When visiting a new mini-batch we compute the gradient using
\eqref{eq:MCELBO}.  For all previously visited mini-batches we compute
the importance weights and modify the gradient according to
\eqref{eq:impgrad}.  The whole gradient is formed by summing up the
terms for all mini-batches. It is important to note that the
importance sampling changes the weight of the gradient, decreasing it
towards zero for the mini-batches evaluated under clearly different
parameter settings, and transforms the gradient to better match one
that would have been calculated under the current approximation.

This algorithm provides a justified version of SAG for automatic
variational inference. The computational cost is higher than for
standard SAG since we need to evaluate the importance weights and
compute the terms related to the gradient of the transformation for
all past mini-batches.  There is, however, no additional memory
overhead and the amount of evaluations for the gradient of the model
itself is the same. This overhead for importance sampling the
gradients for other batches is not negligible, but usually still small
enough that the resulting algorithm outperforms a naive implementation
of SAG because of vastly more accurate gradient estimates, as shown in
Section \ref{sec:isage}. In case updating the past gradients becomes
too costly, a simple remedy is to use only the latest $K$ mini-batches
for some reasonable choice of $K$.

\section{EXPERIMENTS}
In this section we first demonstrate how the behavior of importance
sampling depends on the dimensionality of the approximation. We then
empirically compared both I-SGD and I-SAG for real variational
inference tasks on a range of alternative models and settings.

\subsection{DIMENSIONALITY OF THE APPROXIMATION}
\label{sec:dimensionality}

\begin{figure}[t!]
  \centering
  \includegraphics[scale = 0.53]{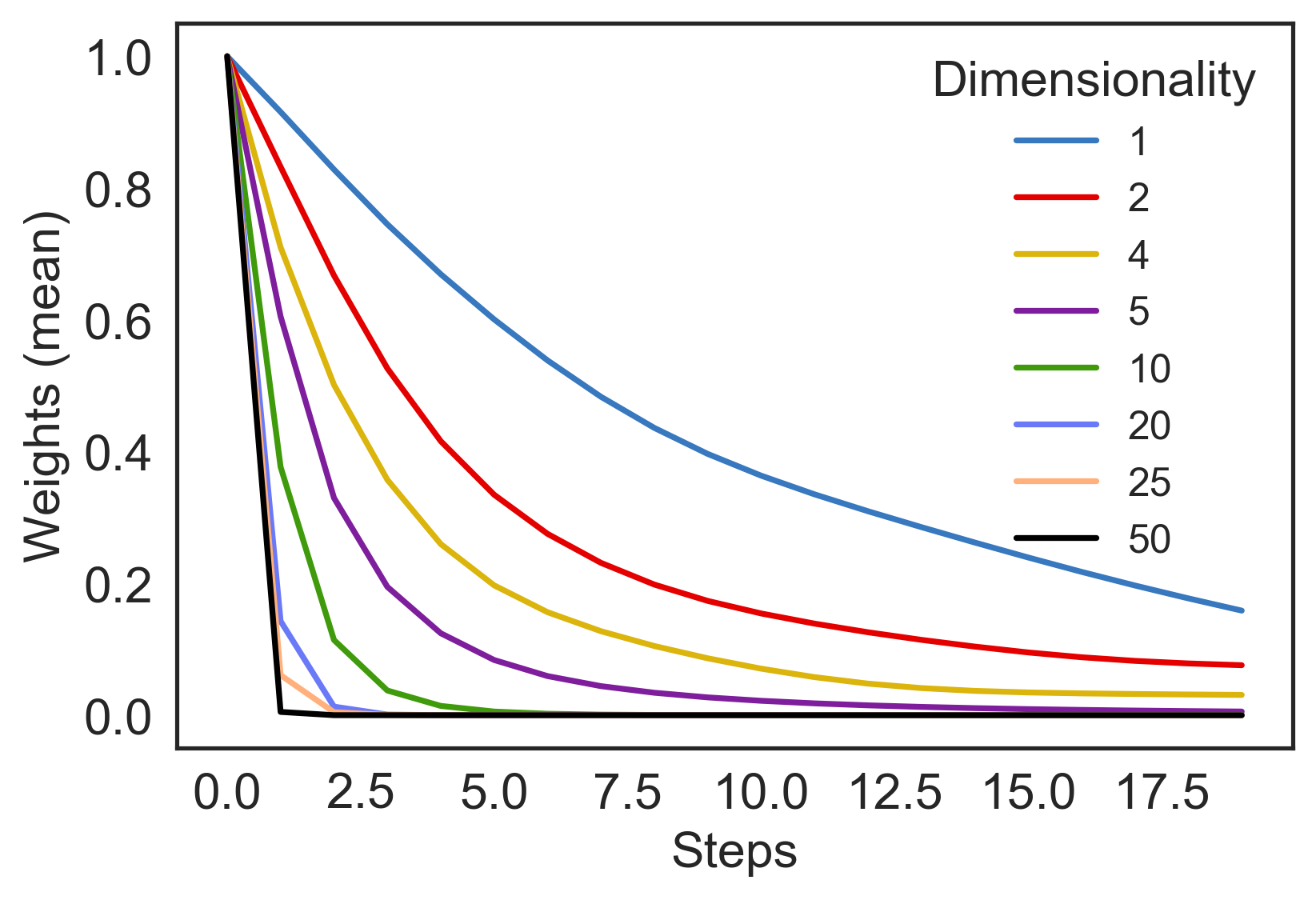}
  \caption{\textbf{Importance sampling and dimensionality}:
    Illustration of the behavior of the importance sampling weights
    during optimization for approximations of varying dimensionality.
    The approximation $q(\theta|\lambda)$ is over a 100-dimensional
    space and is factorized as $\prod_i q(\theta_{S_i}|\lambda_i)$ so
    that the color of the line indicates the size of each term; for
    example, the purple line corresponds to factorized approximation
    with 20 terms of 5 variables.  For fully factorized approximation
    (blue line) and still for factors of 5-10 dimensions the weights
    decay slowly and it pays off to take a few gradient steps before
    re-calculating the gradient.  For factors of 25 (yellow line) or
    more dimensions the weights drop to zero practically
    immediately. The I-SGD algorithm still works, but does not give
    any speed advantage.}
  \label{fig:weight-decay}
\end{figure}

Figure \ref{fig:weight-decay} studies importance weights of
approximations factorized at different granularities on a
100-dimensional diagonal multivariate Gaussian. An important
observation is that even if importance sampling itself fails for
factors of high dimensionality, the I-SGD algorithm degrades
gracefully. For low-dimensional factors, up to at least 5-10
dimensions, we can safely take 5-10 steps with each mini-batch while
still having accurate gradient estimates.  When the dimensionality of
individual factors reaches around $25$ the weights tend to zero
already after a single gradient step, but the algorithm does not break
down. It merely needs to proceed immediately to the next step,
reverting back to standard SGD.
\begin{figure*}[t!]
  \centering
  \begin{subfigure}[t]{.28\textwidth}
    \includegraphics[scale = .5]{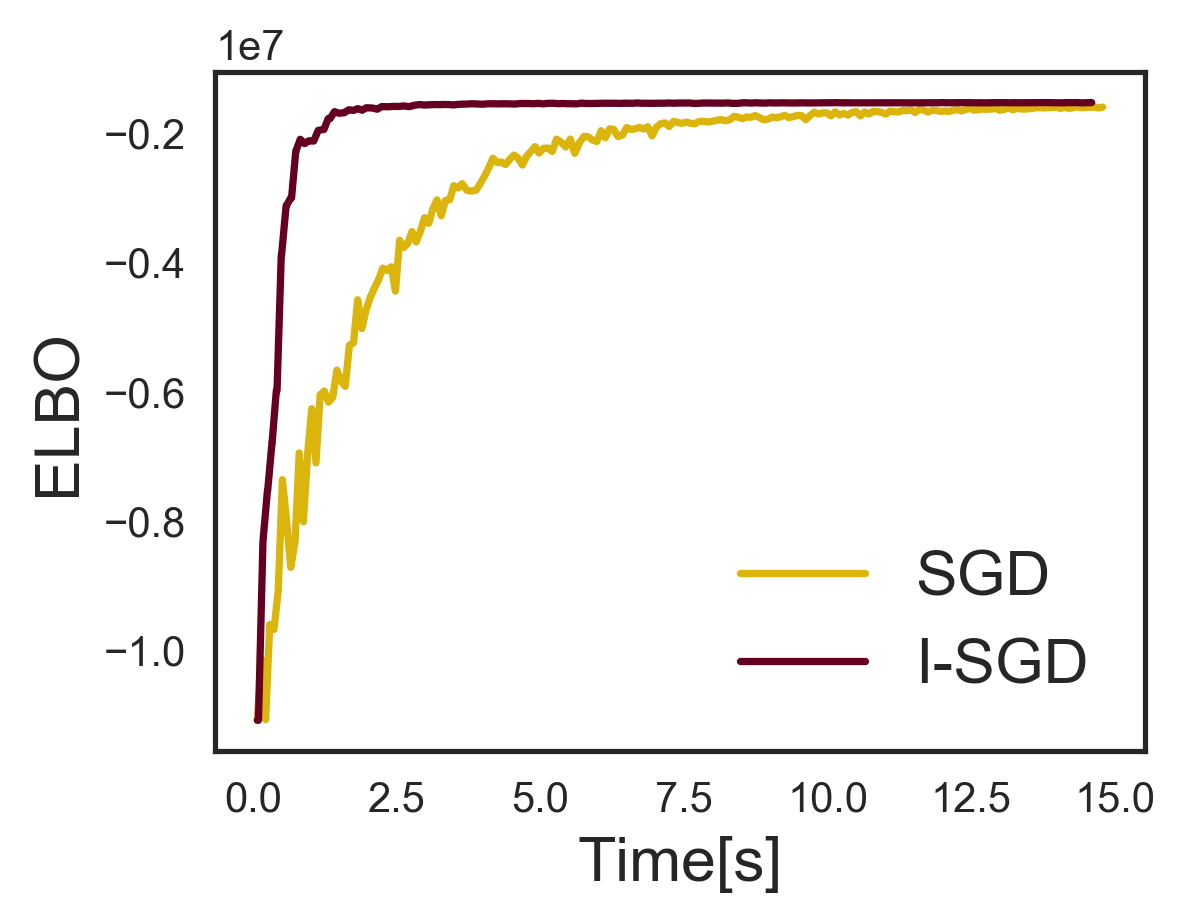}
    \caption{}
  \end{subfigure}
  \begin{subfigure}[t]{.28\textwidth}
    \includegraphics[scale = .5]{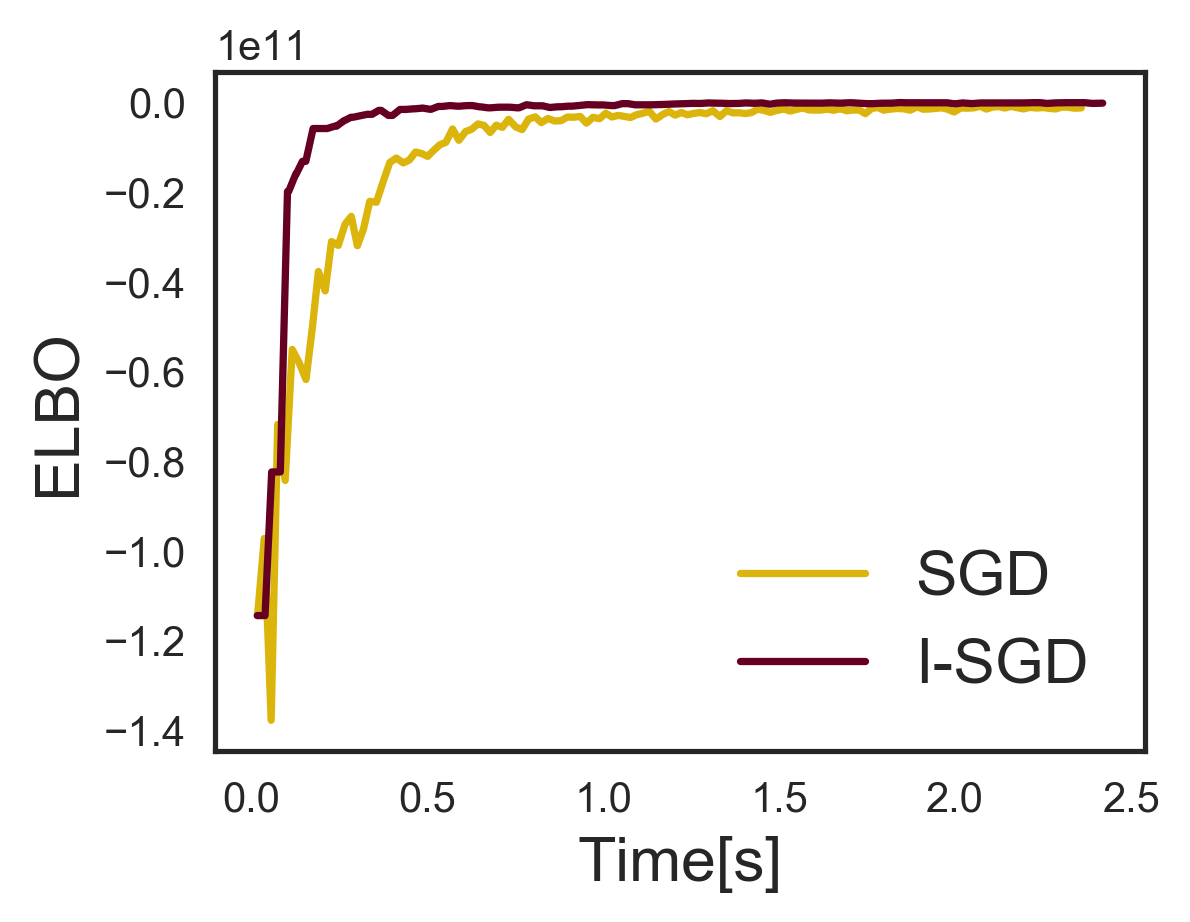}
    \caption{}
  \end{subfigure}
  \begin{subfigure}[t]{.28\textwidth}
    \includegraphics[scale = .5]{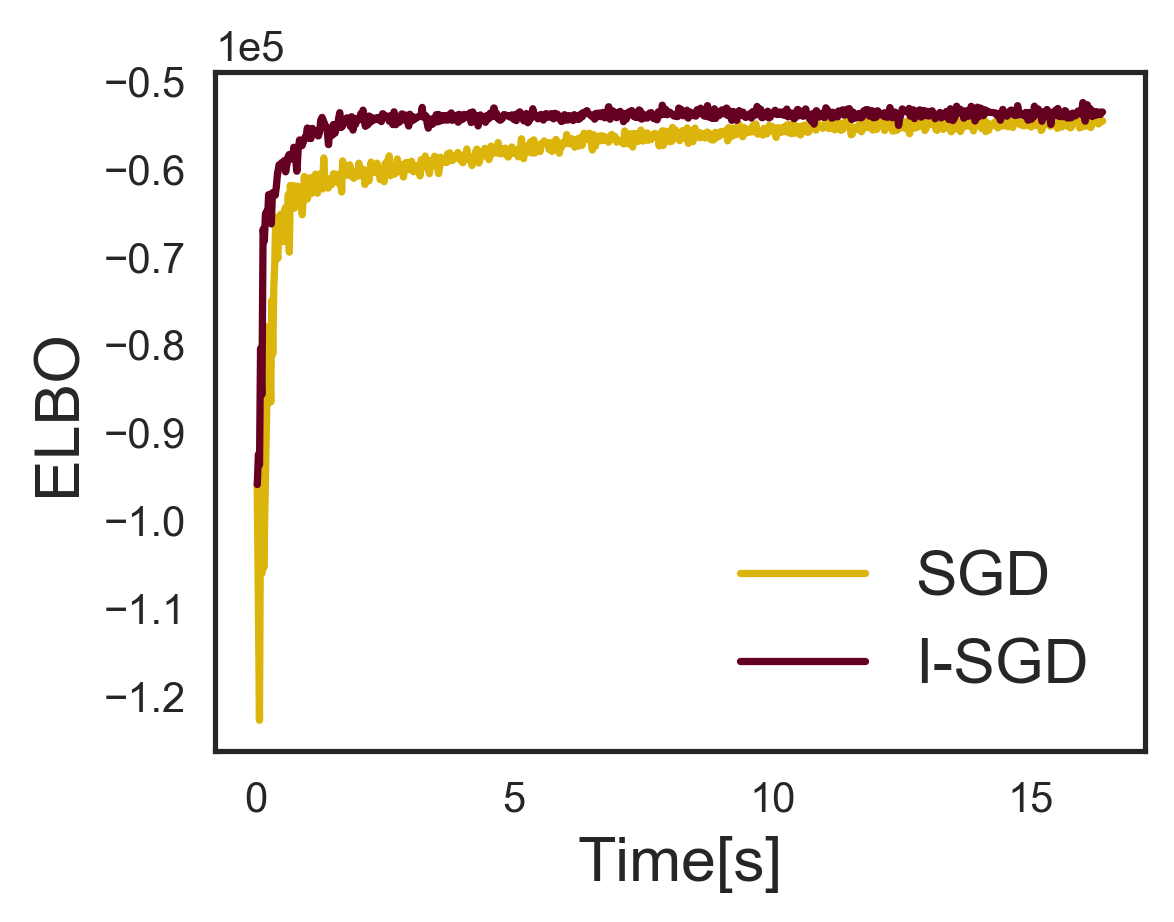}
    \caption{}
  \end{subfigure}
  \begin{subfigure}[t]{.28\textwidth}
    \includegraphics[scale = .5]{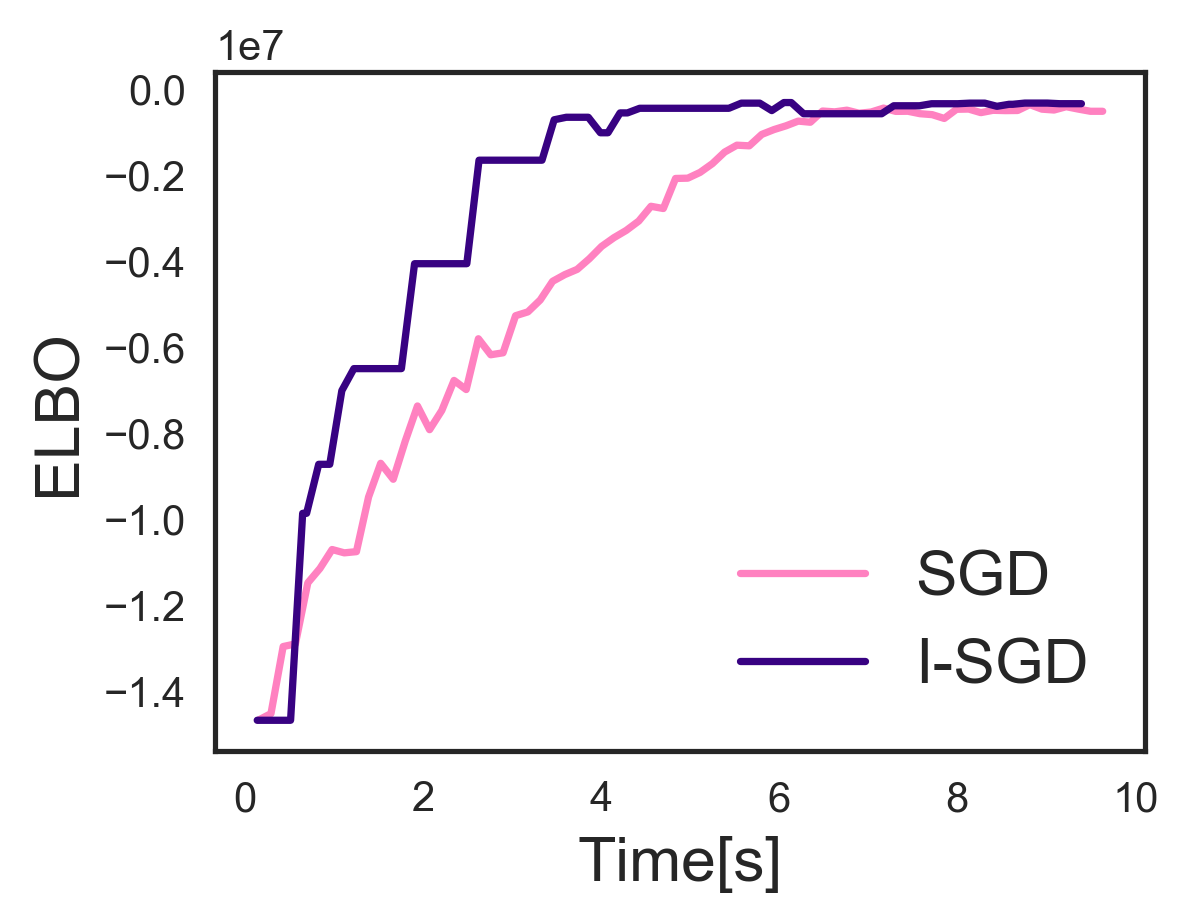}
    \caption{}
  \end{subfigure}
  \begin{subfigure}[t]{.28\textwidth}
    \includegraphics[scale = .5]{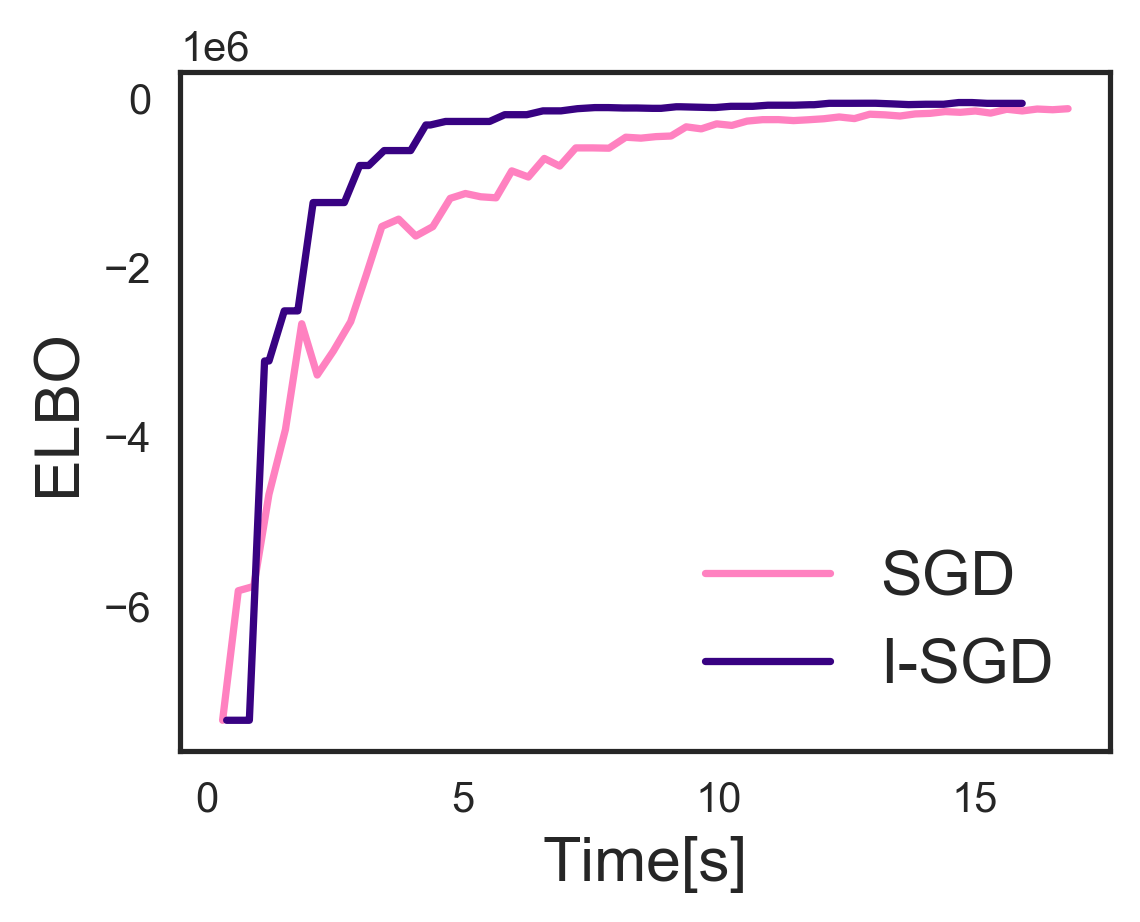}
    \caption{}
  \end{subfigure}
  \begin{subfigure}[t]{.28\textwidth}
    \includegraphics[scale = .5]{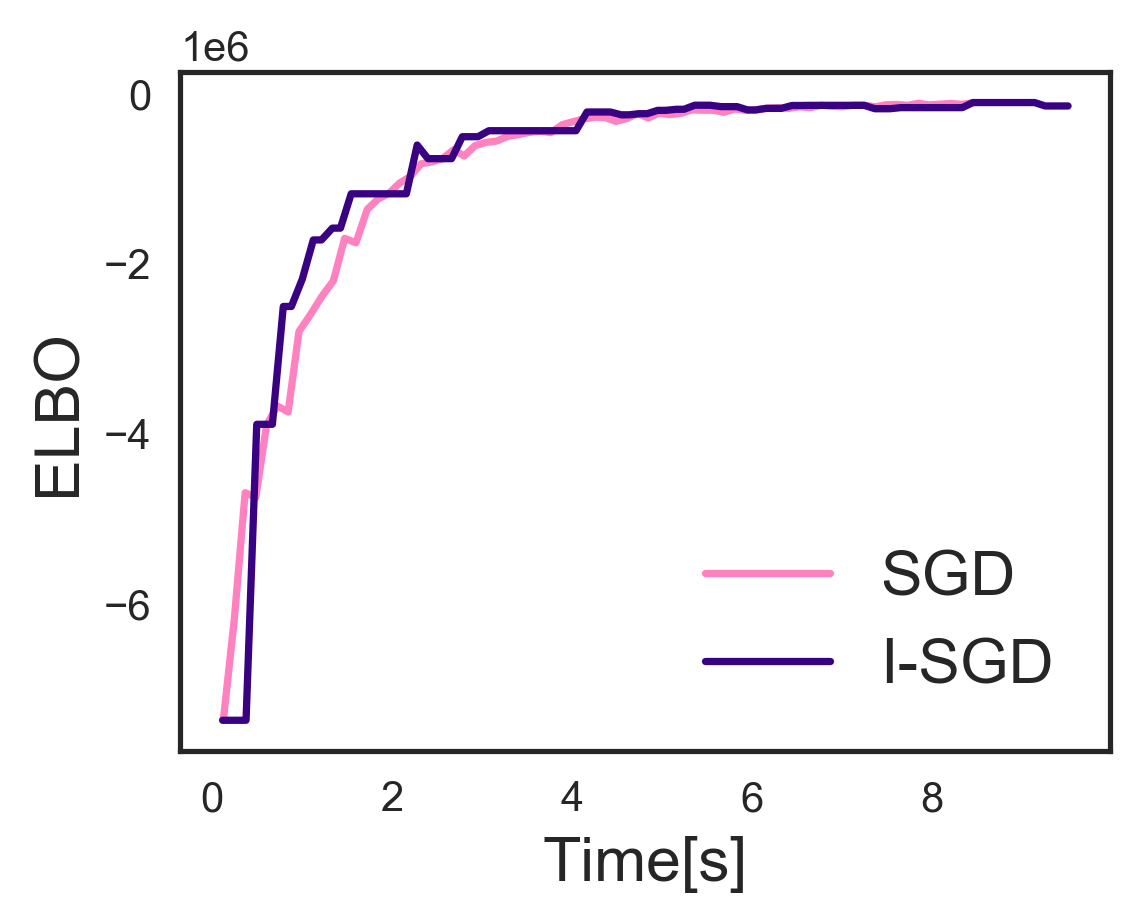}
    \caption{}
  \end{subfigure}
  \caption{\textbf{Performance of I-SGD on different models and
      gradient estimates}: We evaluate the performance of I-SGD for
    reparameterization gradients in the top row and for score function
    gradients on the bottom row. For the reparameterization gradients
    our algorithm converges an order of magnitude faster than SGD for
    all three models.  The computational savings in score function
    gradient is directly proportional to the complexity of the model's
    log-probability and the simplicity of the variational
    approximation; in (d) evaluating the model dominates the total
    cost whereas in (f) the advantage of I-SGD is lost because less
    computation is saved.  For description of the models refer to
    Section~\ref{sec:impsamsgd}.}
  \label{fig:experiments}
\end{figure*}
\begin{figure}
  \centering
  \includegraphics[scale = 0.6]{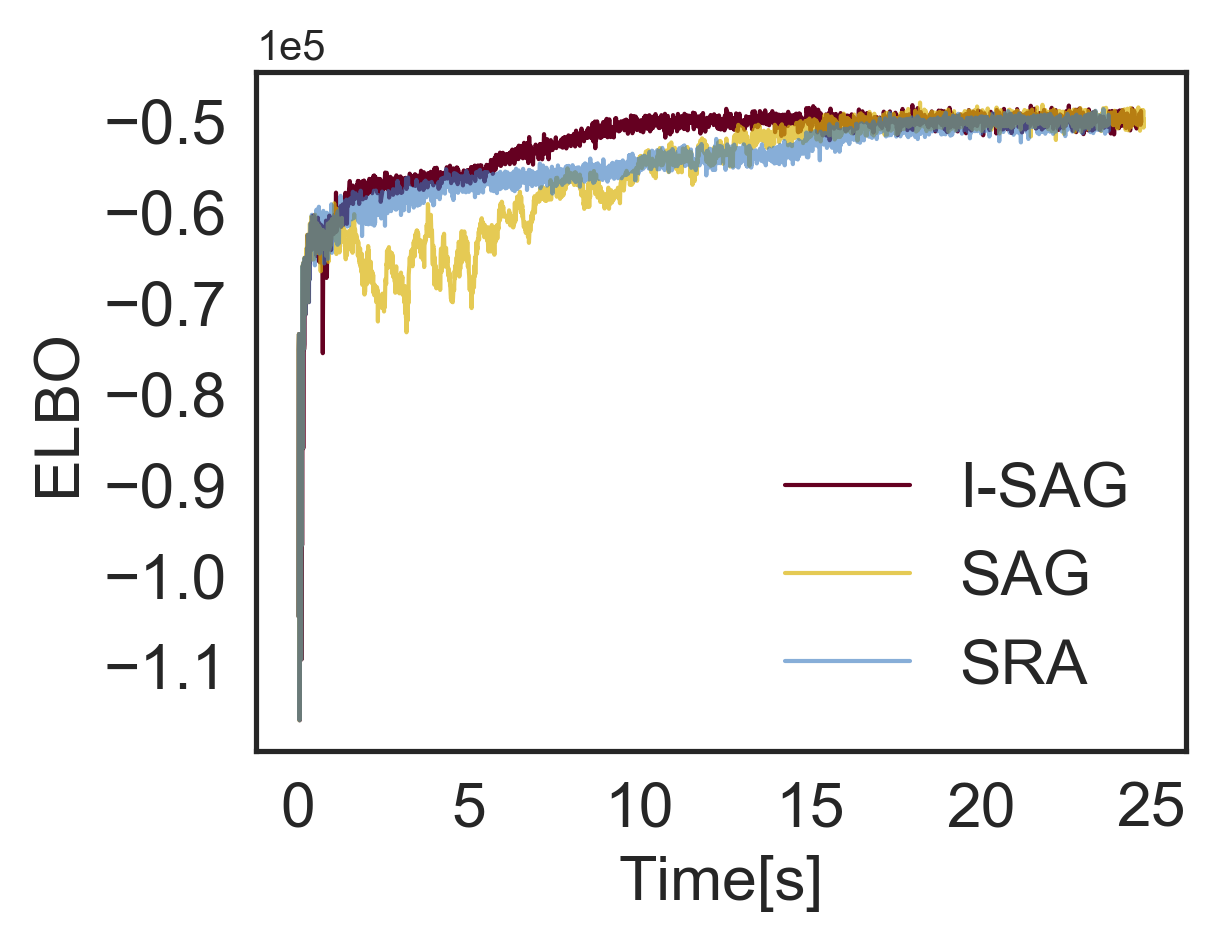}
  \caption{\textbf{Importance Sampled Stochastic Average Gradient}:
    Standard SAG algorithm (yellow line) performs poorly when using
    Monte Carlo approximation for the gradients. This can be in part
    overcome by considering a running average (SRA) of the mini-batch
    gradients instead (blue line, with $\alpha=0.9$ as the decay
    factor), letting the algorithm ignore stale gradients. The I-SAG
    algorithm (red line) outperforms both because it down-weights
    mini-batch gradients evaluated earlier in a more justified manner
    and because it also transforms the historical gradients to match
    the current approximation instead of using them directly.}
  \label{fig:sag}
\end{figure}
\subsection{IMPORTANCE SAMPLED SGD}
\label{sec:impsamsgd}
We show in Figure~\ref{fig:experiments} how I-SGD consistently bests
SGD for a variety of different models, especially for the
reparameterization estimate.  For these experiments we use fully
factorized mean-field approximation,

We first apply the reparameterization estimate for three probabilistic
models: (a) a diagonal multivariate Gaussian with a standard normal
prior on the mean and Gamma priors on the precisions, trained on $N =
50000$ data points of $D = 500$ dimensions; (b) a Bayesian linear
regression model with standard normal priors on the weights and a
Gamma prior on the precision with $N = 50000$ and $D = 500$; and (c) a
Gaussian mixture model with the usual conjugate priors, $N = 10000$,
$D = 2$ and $K = 25$ clusters.  We used fully factorized variational
approximations for all models, with $\Softplus$ transforms for the
positively constrained precision parameters and the stick breaking
transform for the mixture weights of the last model. For all three
choices the I-SGD algorithm with $t=0.9$ converges to the same optimal
solution as SGD, but does so in roughly an order of magnitude
faster. For fair comparison the size of the mini-batch, the initial
learning rate were chosen for each method to work well for SGD,
forcing I-SGD to use the same choices.  For both algorithms, we used M
= 1 sample to estimate the gradients and Adam \citep{adam} to
adaptively control the learning rate during optimization.

We then compare I-SGD and SGD using the score function estimates on a
Poisson likelihood with a single Gamma prior on the rate (d), and
Bayesian linear regression models with large (e) and small (f)
mini-batch. The variational approximations used were the same as the
priors. For the sub-plots (d) and (e) evaluating the log-probability
takes long compared to evaluating the gradient of the approximation
because of a large mini-batch size, and hence I-SGD is faster. With a
smaller mini-batch size (f) the advantage is lost because evaluating
the gradient of the approximation starts to dominate. We used $M =
100$ samples and did not consider variance reduction techniques for
simplicity.

\subsection{IMPORTANCE SAMPLED SAG}
\label{sec:isage}

Figure~\ref{fig:sag} compares the I-SAG algorithm
(Algorithm~\ref{alg:isag}) against naive implementation of SAG
\citep{sage}. Both algorithms are initialized by passing once through
the data with I-SGD, to provide the initial estimate for the full
batch gradient.

While SAG eventually converges to the right solution, the progress is
slow and erratic due to stale mini-batch gradients being accumulated
into the full gradient. I-SAG fixes the issue by not only
down-weighting gradients corresponding to mini-batches visited several
updates ago, but also by transforming the gradients to match the
current approximation. The additional computation required for
adapting the gradients for other mini-batches results in a
computational overhead of, here, roughly 30\% per iteration, but the
improved accuracy of the batch gradient estimate is more than enough
to overcome this.

Stochastic running average (SRA) provides another baseline that
down-weights older mini-batches exponentially. Similar to I-SAG, it
avoids using mini-batches with badly outdated gradient estimates, by
using a simple weighting scheme without transforming the gradients.
It outperforms SAG, but converges more slowly than I-SAG. Hence, I-SAG
is stable implementation of SAG for variational inference,
outperforming the alternative of running averages often considered as
a remedy for the issues of SAG.

\section{DISCUSSION}
Automatic variational inference using automatically differentiated
gradients has in recent years become a feasible technique for
inference for a wide class of probabilistic models, extending the
scope of variational approximations beyond simple conjugate models
towards practical probabilistic inference engines. While standard
computational platforms and advances in convex optimization are
readily applicable for gradient-based variational inference, the need
to use Monte Carlo approximation to estimate the gradients necessarily
induces a computational overhead -- with very few samples the
gradients are noisy whereas the cost grows linearly as a function of
the samples.

Our work addressed this central element, discussing ways to speed up
the gradient-based inference of variational approximations. By
highlighting how the gradient computation separates into two steps we
derived an importance-sampling estimate for the gradient that often
only needs to evaluate the computationally cheaper part to provide the
estimate. Skipping the computationally costly evaluation of the
gradient of the model itself as often as possible lead to a practical
speedup that is independent of other improvements provided by more
advanced optimization algorithms \citep{svrg, saga}.  Our method
relies on the inverse transformation being unique and efficient to
compute.  This might not be the case for complex structured
approximations or approximations parameterized by neural networks; we
leave more efficient extensions for such cases as future work.

We demonstrated the core idea in creating a more efficient stochastic
gradient descent algorithm for both reparameterization \citep{dsvi,
  advi} and score function \citep{bbvi} estimates used for variational
inference.  In addition, we formulated a theoretically justified
variant of stochastic average gradients \citep{sage} applicable for
variational inference. The idea, however, extends well beyond these
special cases. For example, the rejection sampling variational
inference \citep{naesseth} can be readily combined with our importance
sampling strategy and is expected to result in a similar speedup.

Our main focus was in practically applicable algorithms, with much of
the theoretical analysis left for future work.  Two particular
directions are immediately apparent: (a) The decision of when to use
importance sampling estimates and (b) the behavior for approximations
that do not factorize into reasonably small factors.  In this work we
showed how simple randomized procedure for determining whether to
re-compute the gradient for a new mini-batch results in practical and
robust algorithm, but more theoretically justified decisions such as
inspecting for example the variance of the importance sampling
estimate could be considered. The proposed algorithms are efficient
for approximating factors of dimensionality up to roughly ten; for
factors of higher dimensionality the algorithms revert back to the
standard variants since all gradients need to be computed from scratch
for every iteration.

\subsubsection*{Acknowledgements}
This work was financed by the Academy of Finland (decision number
266969) and by the \emph{Scalable Probabilistic Analytics} project of
Tekes, the Finnish funding agency for innovation.

\bibliographystyle{plainnat} \bibliography{paper}

\end{document}